\documentclass[letterpaper, 10 pt, conference]{ieeeconf}  

\IEEEoverridecommandlockouts                              

\usepackage{graphicx} 
\usepackage{times} 
\usepackage{microtype}
\usepackage{amsmath} 
\usepackage{amssymb}  

\usepackage[numbers]{natbib}

\usepackage[bookmarks=true]{hyperref}
\hypersetup{pdfborder = 0 0 0}

\usepackage{caption}
\usepackage{todonotes}
\usepackage{xcolor}

\begin{document}

\title{\LARGE \bf
	Virtual-to-Real-World Transfer Learning for Robots on Wilderness Trails\\
}

\author{Michael L. Iuzzolino$^{1}$ and Michael E. Walker$^{2}$ and Daniel Szafir$^{3}$
	\noindent \thanks{$^{1}$Department of Computer Science, University of Colorado Boulder.
		{\tt\small michael.iuzzolino@colorado.edu}}%
	\thanks{$^{2}$Department of Computer Science, University of Colorado Boulder.
		{\tt\small michael.walker-1@colorado.edu}}%
	\thanks{$^{3}$Department of Computer Science and ATLAS Institute, University of Colorado Boulder.
		{\tt\small daniel.szafir@colorado.edu}}%
}

\maketitle


\begin{abstract}
	Robots hold promise in many scenarios involving outdoor use, such as search-and-rescue, wildlife management, and collecting data to improve environment, climate, and weather forecasting. However, autonomous navigation of outdoor trails remains a challenging problem. Recent work has sought to address this issue using deep learning. Although this approach has achieved state-of-the-art results, the deep learning paradigm may be limited due to a reliance on large amounts of annotated training data. Collecting and curating training datasets may not be feasible or practical in many situations, especially as trail conditions may change due to seasonal weather variations, storms, and natural erosion. In this paper, we explore an approach to address this issue through virtual-to-real-world transfer learning using a variety of deep learning models trained to classify the direction of a trail in an image. Our approach utilizes synthetic data gathered from virtual environments for model training, bypassing the need to collect a large amount of real images of the outdoors. We validate our approach in three main ways. First, we demonstrate that our models achieve classification accuracies upwards of 95\% on our synthetic data set. Next, we utilize our classification models in the control system of a simulated robot to demonstrate feasibility. Finally, we evaluate our models on real-world trail data and demonstrate the potential of virtual-to-real-world transfer learning.
\end{abstract}

\section{INTRODUCTION}
Robots have shown significant aptitude in data-gathering and inspection tasks across a variety of domains. Aerial robots are especially adept at such tasks (see \cite{Szafir2016} for a survey). Due to substantial cognitive demands, many aerial robot systems are manned by teams of two humans, one acting as pilot and the other as mission specialist orchestrating the collection and analysis of data \cite{peschel2013human}. Extending autonomous navigation capabilities could positively impact robot operations by attenuating cognitive demands and allowing human operators to focus on other mission-critical tasks that involve high-level decision making.

Autonomous navigation of outdoor trails presents a complex, non-trivial perception and planning problem. Unlike well-defined environments, such as roadways and sidewalks in urban areas, wilderness trails consist of drastically varying features (e.g., gravel path, game trail, backcountry dirt road), traverse highly variable terrains, and span vastly differing biomes (e.g., forests, meadows, mountains), all under various seasonal and lighting conditions. Dense vegitation and large obsticles may pose significant visibility constraints, while GPS measurements may be unreliable or even unavailable \cite{hsieh2007adaptive}. Consequently, autonomous navigation of unknown terrain and environments is an active area of research within the fields of machine learning and robotics.

Deep learning approaches are establishing state-of-the-art results for robot perception, planning, navigation tasks. However, such approaches require large, labeled training datasets that often require exhaustive human labor for collection and labeling. In many instances, collecting and labeling these datasets poses significant challenges, some of which may be insurmountable due to logistical issues. For example, search-and-rescue is particularly critical during harsh weather conditions, but it is these hazardous conditions in which it is most difficult to collect training data for data-driven approaches, such as deep learning.

In this paper, we demonstrate a deep learning approach that may mitigate these issues through the utilization of transfer learning between virtual and real-world domains. We propose a solution for training neural networks on synthetic images of virtual outdoor trails, where a neural network learns to identify the direction of the trail within an image, and demonstrate that the features learned on the virtual dataset are capable of transferring to real-world domains for trail perception. Our method alleviates the need for exhaustive real-world data collection and laborious data labeling efforts.

\section{Related Work}
Our approach draws from the fields of robotic perception, computer vision, and deep learning. Below, we discuss image classification and object detection for trail perception. We then review advances in transfer learning between real-world datasets and discuss extensions of transfer learning to virtual and real-world datasets for use in robot perception and navigation tasks.

Previous efforts to solve the problem of autonomous pathfinding and navigation focused on trail segmentation using low-level features such as image saliency or appearance contrast \cite{peschel2013human, giusti2016machine}. However, more recent approaches have leveraged deep learning to produce cutting-edge results for elements of robotic navigation, such as trail perception and object detection. In the work of \citet{giusti2016machine}, a hiker was equipped with three head-mounted GoPro cameras with left-, center-, and right-facing orientations and traversed alpine regions of Switzerland for 8 hours, resulting in a dataset of 24,747 natural trail images. The camera setup allowed for automatically labeled data: images collected by the left-facing GoPro camera were labeled as \textit{left}, and so on.

This dataset was used to train a convolutional neural network that learned to discriminate on salient features that best predict the most likely classification of the image. This method achieved classification accuracies of 85.2\% and outperformed conventional computer vision methods, such as hue-based saliency mapping for RBF kernel SVM classification (52.3\%), and is comparable the performance of humans (86.5\%) \cite{giusti2016machine}. The network was qualitatively evaluated as a control system for a real-world aerial drone with a monocular camera and demonstrated moderate success.

While \citet{giusti2016machine} demonstrated promising results, their approach relies on real-world data collection and may thus be limited due to issues arising from battery-life, human fatigue, data collection errors due to incorrect head orientation and mislabeling of data, or seasonal availability and safety. In addition, this approach may not extend to inaccessible, novel, and/or dangerous environments, such as rugged winter trails or extraterrestrial environments (e.g., for use in robotic space exploration on Mars or the Lunar surface). A possible solution to these challenges is transfer learning, an active area of research within the deep learning community, where knowledge representations are learned in one domain and utilized to accelerate learning in a related domain. For
instance, research has revealed that convolutional neural networks trained on natural images learn generalizable features, such as Gabor filters and color blobs \cite{yosinski2014transferable}, that form the basis of many datasets, such ImageNet \cite{ILSVRC15} datasets.

Our approach is inspired by transfer learning; however, instead of transferring from one real-world domain to another, we are interested in the notion of transferring knowledge learned in virtual environments to the real world. For example, prior work has developed a mapless motion planner for real environments by training a deep reinforcement model in synthetic settings \cite{tai2017virtual}. After training in a well-defined simulation, the system converges upon an optimal set of navigational policies that are then transferred to a real-world robot capable of navigating a room with static obstacles. This work highlights the potential of virtual-to-real transfer learning in domains where a well-defined simulation is available. However, this work does not address the challenges of perception and navigation in complex environments where simulations may be lacking or non-existent. Our work in this paper further explores the potential of virtual-to-real-world transfer learning to address the challenges raised by complex domains, such as wilderness trails.

\begin{figure}
	\centering
	\includegraphics[width=1\columnwidth]{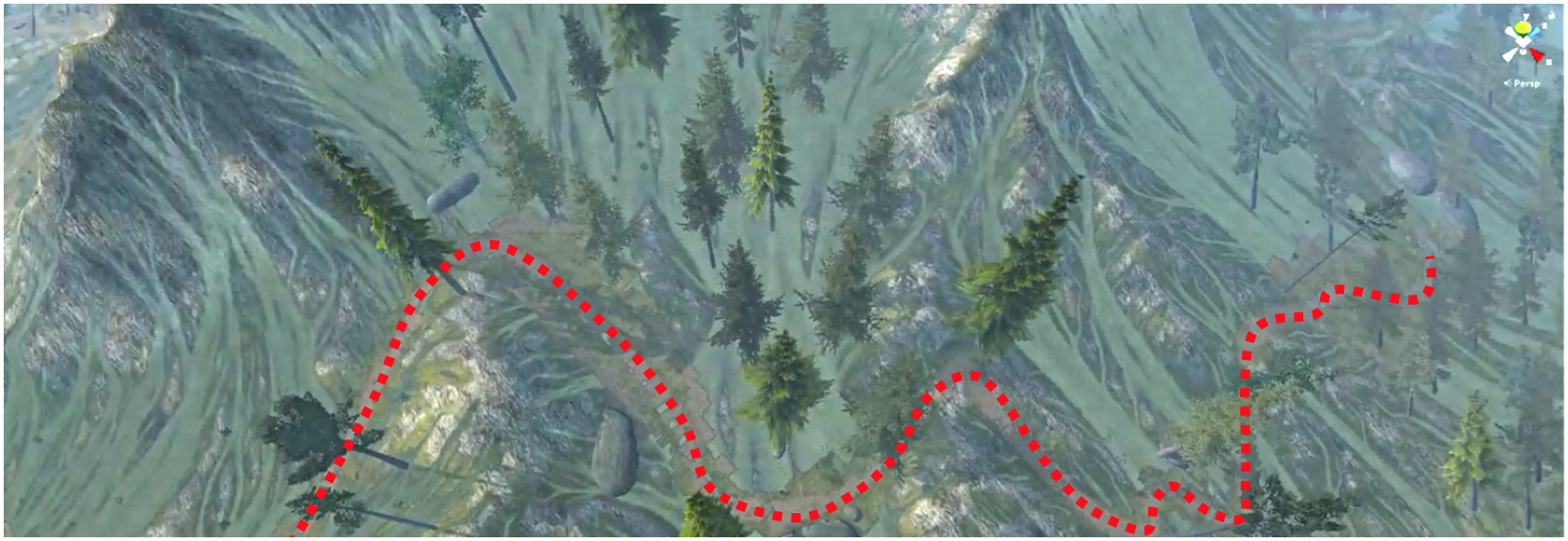}
	\caption{Birds-eye view subsection of trails (dotted-red line) traveled by virtual camera and robot.}
	\label{fig:paths}
\end{figure}

\section{Approach}
\label{sec:Approach}
To explore the concept of virtual and real-world trail navigation, we created a virtual environment for synthetic data collection. Below, we discuss the details of the virtual environment. Then we outline our methods for data collection, processing, and constructing three different archetypal neural networks. Finally, the last two sections describe the integration of the trained neural network with the Unity environment---a cross-platform 3D game engine---and the method used for evaluating our models on real-world data to demonstrate virtual-to-real-world transfer learning.

\subsection{Virtual Environment}
\label{sec:unity_enviornment}
To create our virtual environment, we used Unity, a game and animation engine for developing virtual interactive 3D environments. Using the built-in terrain editor and readily available 3D models of natural objects (trees, rocks, grass, etc.) from the Unity Asset Store, we assembled a virtual scene of an alpine mountain with a web of dirt trails spanning the landscape (Figure \ref{fig:paths}). The paths in the environment held many similarities to real-world trails: they branched, curved around rocky corners and wooded areas, changed elevation, and contained ambiguous trail sections.

\subsection{Data Collection}
\begin{figure}
	\centering
	\includegraphics[width=1.0\columnwidth]{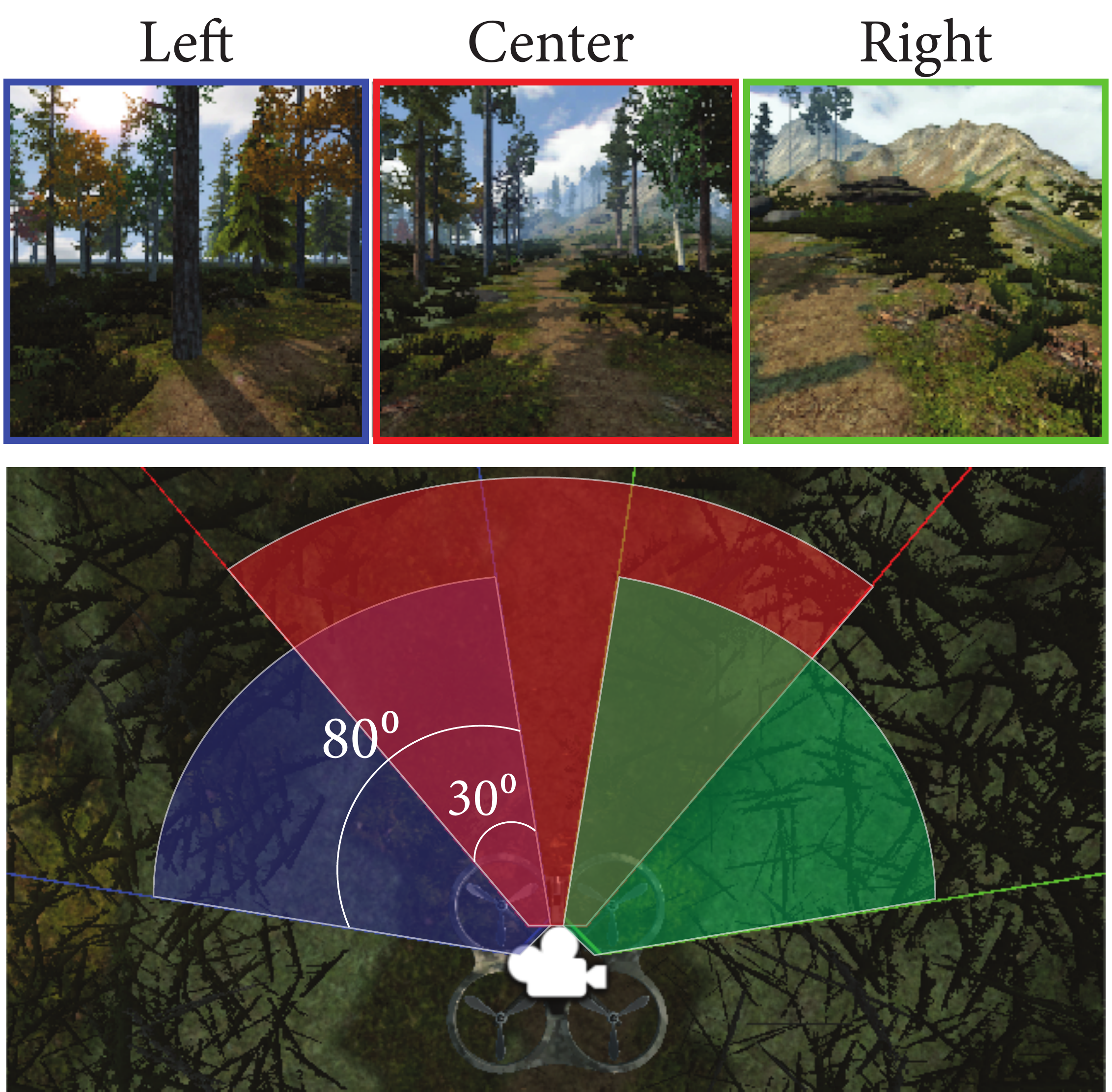}
	\caption{Top-down view of camera configuration for data capture in Unity environment.}
	\label{fig:processedimages}
\end{figure}

A single path in the Unity environment was randomly chosen and utilized for all data collection. A virtual robot was placed onto the path and a C\# control script was attached to the robot that enabled it to deterministically traverse the path multiple times. Three cameras, each with $400 \times 400$ pixel resolution, 30 frames per second (FPS), and an 80$^{\circ}$ field-of-view (FOV) were attached to the robot. The combined FOV capture for all three cameras spanned 180$^{\circ}$ with each periphery camera having 30$^{\circ}$ of overlap with the central camera (Figure \ref{fig:processedimages}). The camera configurations were determined in conjunction with the data capture setup in \cite{giusti2016machine}, GoPro camera design specifications, and the results from preliminary virtual-world data capture trials. The virtual robot's roll and pitch were constrained to 0$^{\circ}$, with the yaw always set to a value that directed the robot toward the center of the path. The robot traversed the path and collected a total of 20,269 images (\textit{center}: 6821, \textit{left}: 6829, \textit{right}: 6619). The screen-shot bundles were labeled as either \textit{center}, \textit{left}, or \textit{right}---depending on which camera the images were captured from---and stored locally.

\subsection{Data Processing}
\label{sec:dataproc}
Standard image processing practices, such as resizing and normalization, were followed. The images were resized to $100 \times 100 \times 3$ pixels. This allowed for faster processing and lower memory consumption, which is especially problematic for large neural network models. We then normalized the images to account for the highly variable range of pixel values. Non-normalized data is problematic during back-propagation for most machine learning algorithms, where weight changes are computed by the accumulation of the gradient, multiplied by a scalar learning rate. With non-normalized feature vectors, the result is typically an oscillatory behavior of the gradients, as the weights of some features are over-corrected whereas others are under-corrected. Consequently, we normalized the pixel-space of our images to values between $[0, 1]$. To normalization across a large, high-dimensional dataset, we opted to normalize on each image and per color channel, rather than across the distribution of all images in the dataset.

\noindent\textit{Training, Validation, and Test Sets:}
The virtual data collected via Unity was split into three sets: training, validation, and testing. The real-world dataset from \cite{giusti2016machine} was utilized as an additional test set to demonstrate the transferability of features between virtual and real-world domains. The splits and distributions are presented in Table \ref{table:data_dist}.

\subsection{Model Architectures}
We explored three different model architectures: standard feed-forward deep neural networks (DNN's), convolutional neural networks (CNN's), and recurrent neural networks (RNN's). In the following subsections, we outline the models' hyperparameters, and input/output structures.

\subsubsection{Deep Neural Network}
\begin{figure*}
	\centering
	\includegraphics[width=1\textwidth]{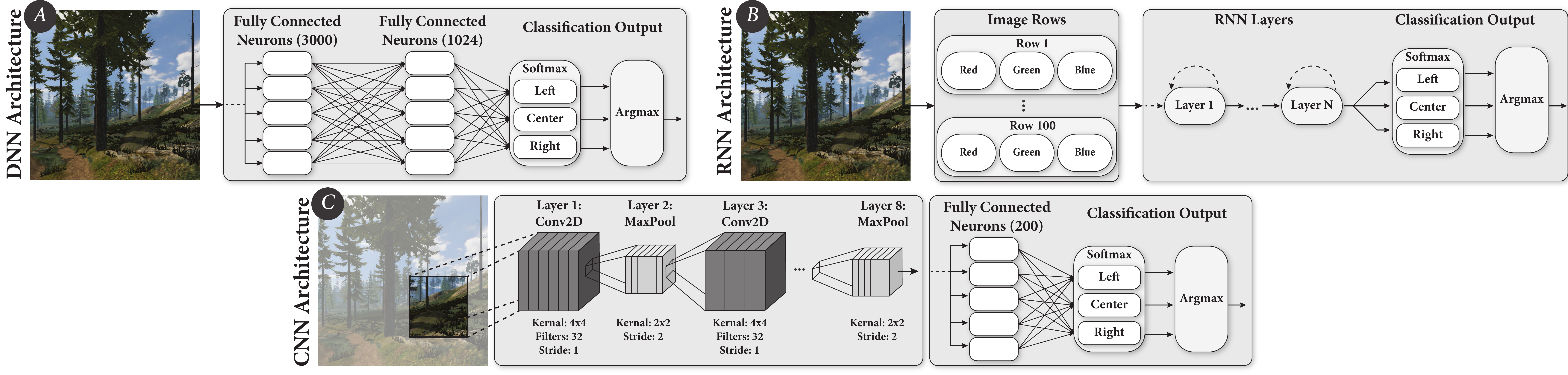}
	\caption{\textnormal{We explored DNN (A), RNN (B), and CNN (C) models for classifying virtual trail imagery.}}
	\label{fig:architectures}
\end{figure*}

The feed-forward network is outlined in Figure \ref{fig:architectures} (A). The $100 \times 100 \times 3$ pixel input images are flattened to a $30000 \times 1$ dimensional input vector and fed to the input layer of the DNN, which contains $100 \times 100 \times 3 = 30000$ input neurons. This architecture implemented three hidden layers (not shown in Figure \ref{fig:architectures}) and utilized rectified linear unit (ReLU) activation functions. The output of the last hidden last layer is sent to a final output layer that consists of three neurons, where each maps to a corresponding classification prediction of left, center, or right. A \textit{softmax} activation is applied to the outputs, establishing a proper probability distribution over which the \textit{argmax} yields the classification prediction.

\subsubsection{Convolutional Neural Network}
The architecture of this model replicates that of the CNN utilized in \cite{giusti2016machine} (Figure \ref{fig:architectures}, C). The $100 \times 100 \times 3$ pixel input images are fed into the first convolution layer, which contains 32 filters, $4 \times 4$ kernels, and a stride of 1. The convolution layer is activated by a sigmoid function and then fed to a max pool layer with kernel sizes of $2 \times 2$ and strides of 2. This block of convolution, activation, and max-pooling is repeated with each unit containing identical parameters a total of four times. The 4th max pooling layer is flattened and fed to a fully connected layer with 200 neurons, and the sigmoid activated output is fed to the final output layer containing three neurons. The output layer is identical to the DNN.

\subsubsection{Recurrent Neural Network}
This architecture is depicted in Figure \ref{fig:architectures} (B). Both Gated Recurrent Units (GRUs) \cite{chung2014empirical} and Long Short Term Memory (LSTM) \cite{hochreiter1997long} cells were explored. The negligible performance difference between the two cell types \cite{jozefowicz2015empirical} prompted us to use the GRU model given its simplicity with respect to the LSTM. We utilized a two-level architecture where each layer contains 32 hidden units. The $100 \times 100 \times 3$ pixel image was reshaped into 100 3-element sequences---one per color channel---where each element consists of 100 values and fed as sequential input into the RNN. The final output layer is identical to the DNN.

\begin{table}
	\vspace{5mm}
	\caption{Dataset Counts and Distribution}
	\centering
	\begin{tabular*}{\columnwidth}{c c c c c}
		\hline\hline
		
		Dataset & Count & Left & Center & Right   \\
		\hline
		Training (Simulated) & 12972 & 33.60\% & 33.85\% & 32.55\% \\
		Validation (Simulated) & 3243 & 35.25\% & 31.85\% & 32.90\% \\
		Test (Simulated) & 4054 & 32.76\% & 34.46\% & 32.78\% \\
		Real Test (Real-World) & 12000 & 33.33\% & 33.33\% & 33.33\% \\
		\hline
		\label{table:data_dist}
	\end{tabular*}
	\vspace{-4.14mm}
\end{table}

\subsection{Training: Loss Functions, Optimizers, and Evaluations}
All models were trained with the same loss function, optimizer, and evaluation metrics. Cross-entropy was used as the loss function and an Adam Optimizer \cite{kingma2014adam} with an initial learning rate of 0.001 was used to minimize the cross-entropy loss. The models were evaluated on their accuracy scores, defined as the ratio between the number of correctly labeled images to the total number of images in the set.

\subsection{Neural Network Integration with Unity}
\label{sec:unity_NN}
In addition to the validation and test accuracy evaluations, and similar to the qualitative evaluation of  \cite{giusti2016machine}, we devised an evaluation within the Unity environment where the neural network was utilized as a controller of a virtual robot. We instantiated the virtual robot onto one of the virtual trails that was not used during training data collection, ensuring it would not see data it had already been trained on. We then allowed the virtual robot to freely explore the environment, and we qualitatively analyzed its behavior at a high level, seeking to observe whether it was able to navigate the trails or deviate from the trail and wander off into the forest.

\begin{figure}
	\centering
	\includegraphics[width=1\columnwidth]{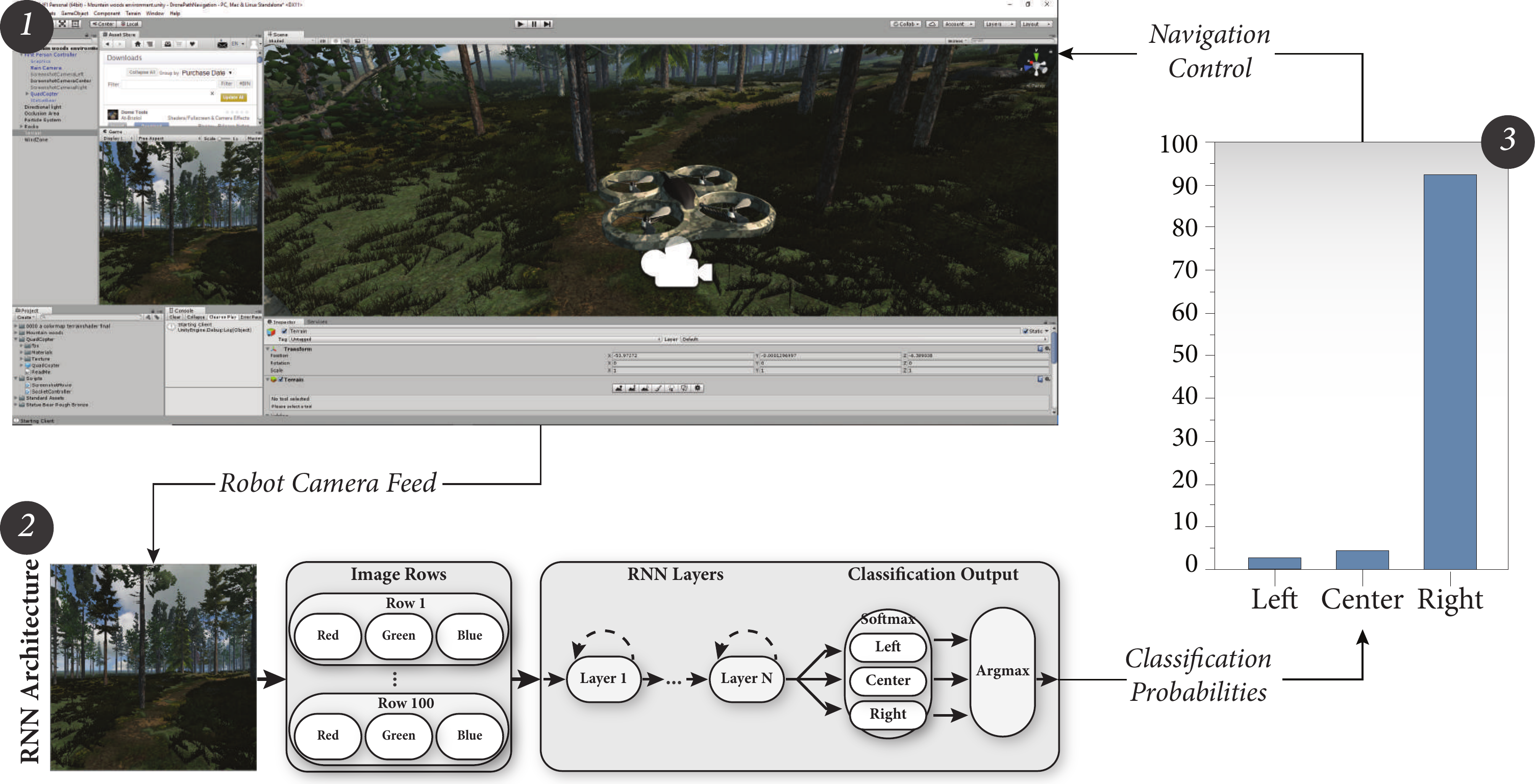}
	\caption{Step 1: A virtual robot placed within the Unity environment captured images for classification by the neural network. Step 2: the neural network receives the image and produces a probability output. Step 3: classification probabilities are visualized, with the resulting command generated by the maximum class probability sent via UDP back into the Unity environment to control the virtual robot's movements.}
	\label{fig:control_system}
\end{figure}

To accomplish this, the Unity scene was adjusted to allow for the direct control of the virtual robot by the neural network's classification predictions. In contrast to the virtual data collection where the three-camera paradigm was utilized, a single camera component was positioned at the center orientation of the virtual robot and set to capture images at 30 FPS. This reflects real-world scenarios where robots typically have a single, forward-facing camera. As soon as the image was captured (Figure \ref{fig:control_system}, Step 1), the neural network processed the image (Figure \ref{fig:control_system}, Step 2) and transmitted the classification prediction via UDP socket back to the Unity scene (Figure \ref{fig:control_system}, Step 3). The UDP packet is then parsed within the Unity environment and the virtual robot then moves deterministically based on the neural network prediction.

The control system was designed as follows: a \textit{center} prediction moves the robot straight ahead, a \textit{left} prediction slows the robot's forward movement and rotates it right, and a \textit{right} prediction slows the robot's forward movement and rotates it left. The classifications on an image corresponds to the source camera orientation. Consequently, images obtained from the left camera during training contained trails on the right-hand side of the image; as a result, when the model predicts \textit{left}, the proper control response is to turn right, toward the direction of the trail. Through this pipeline, the virtual robot was set to navigate the virtual path based solely on the neural network's output of an image taken from a single, forward-facing virtual camera in real time.

\subsection{Evaluation on Real-World Data}
\label{sec:real_world_data}
The real-world dataset from \cite{giusti2016machine} was utilized as a test set on the models trained on virtual data to demonstrate the feasibility of virtual-to-real-world transfer learning. The set was randomly generated by sampling 4,000 images from each classification (left, center, and right), resulting in 12,000 real-world images. This approach guarantees class balance and establishes the test set baseline at 33\% (see Table \ref{table:data_dist}, \textit{Real Test}). The test set images are processed utilizing the methods described in \S \ref{sec:dataproc}, and then fed forward through the virtually trained models to generate a prediction on the real-world image. For every image, the prediction is compared to the image's true label to obtain the accuracy over the set.


\begin{table}[b]
	\caption{Model Results}
	\centering
	\begin{tabular}{c c c c }
		\hline\hline
		Test Set                     & DNN     & CNN     & RNN     \\
		\hline
		Virtual Test Set Accuracy    & 88.70\% & 93.82\% & 95.02\% \\
		\hline
		Real-World Test Set Accuracy & 58.41\% & 38.60\% & 48.51\% \\
		\hline
		\label{table:model_performance}
	\end{tabular}
\end{table}

\section{Results}
\label{sec:Experiments}
All models were trained and evaluated on virtual data ($100 \times 100 \times 3$ pixel images acquired from the Unity environment outlined in \S \ref{sec:unity_enviornment}) for 100 epochs with batch sizes of 128 images. The datasets did not exhibit any significant class imbalance (see Table \ref{table:data_dist}); the predominant class of the three was utilized as the baseline to establish whether the models were achieving better results than a policy of continually guessing the majority class. The baseline for the virtual dataset is 35.25\%, established by the maximum class imbalance from the validation set; the baseline for the real-world dataset is 33.33\%. The models were trained using backpropagation for 50 epochs, which required 1h:23m, 9h:12m, and 2h:4m for the DNN, CNN, and RNN, respectively, on a Macbook Pro with an Intel Iris Pro 1536 MB integrated graphics processor.

\subsection{Virtual Dataset Results}
The RNN provided the best results on the virtual dataset scoring a 95.02\% on test set accuracy, whereas the DNN provided the best results on the real-world datasets, scoring 58.41\%. All three models scored higher than dataset baselines in both virtual and real-world evaluations. The summary of the model performances can be found in Table \ref{table:model_performance}.

\subsection{Unity Follow-Up Evaluations}
As mentioned in \S \ref{sec:unity_NN}, the neural network was integrated with Unity and used as control system for the virtual robot. The RNN model was chosen as the controller due to its top performance on the virtual dataset. In our experiments, we selected sufficiently complex trails---e.g., no straight, level trails---and ensured that the selected trail was not the one used to gather the training data. This ensures that the model is capable of generalizing to novel domains. After selecting an appropriate trail, we placed the virtual robot into the scene and allowed the RNN to govern the autonomous exploration of the environment (see included video submission). Overall, we observed that the robot was largely successful in navigating trails, including those with tight turns and obstacles such as large boulders. Moreover, we observed several instances of ``intelligent" decision-making; in one trial, the robot briefly navigates off the trail after colliding with a large obstruction, but then navigates back to the trail and resumes its travel. While promising, we did observe occasional failures. For example, particular terrain regions that exhibited trail-like features, such as small ridgelines, caused the robot to navigate off the trail and begin following these pseudo-trails features.

\subsection{Real-World Dataset Evaluations}
Real-world evaluation was conduced on 12,000 images from the real-world dataset described in \S \ref{sec:real_world_data}. The DNN, CNN, and RNN models achieved classification accuracies of 58.41\%, 38.60\%, and 48.51\%, respectively. Conventional computer vision approaches, such as hue-based saliency mapping coupled with an RBF kernel SVM classifier trained on the real-world dataset comprising our test set have achieved 52.3\% clasification accuracies \cite{santana2013tracking, giusti2016machine}. Significantly, although none of our models achieved the DNN model or human baseline accuracies from \cite{giusti2016machine}, our work demonstrates that DNNs trained strictly on virtual data can outperform conventional models trained on real-world data.

\section{Discussion}
The experiments on the virtual datasets demonstrate that the deep learning architectures were capable of learning the correct classifications of virtual images, indicated by the high accuracies, ranging from 88.7\% to 95.02\%. These scores strongly exceed the data set baselines of non-intelligently predicting the most frequent class. Importantly, the experiments on the real-world images resulted in classification accuracies ranging from 38.60\% up to 58.41\%, which all exceed the data set baseline of 33.33\%. Interestingly, although the virtually trained models did not outperform the CNN or human baselines for real-world test sets, the DNN did outperform the saliency map / SVM baseline from \cite{giusti2016machine} by more than 6\%. This suggests that virtual-to-real-world transfer learning utilizing deep learning models may outperform conventional computer vision methods for trail perception. Together, these results indicate that discriminating features for perception of real-world trails have been successfully learned exclusively from virtual trails.

We believe there are several ways to further increase the performance of our virtual-to-real world transfer approach. When conducting pilot tests to iterate over potential network models, we found that longer training periods often ended up reducing real-world test accuracy, suggesting that the models are overfitting on the virtual datasets and would benefit from regularization and training on larger datasets. We suspect that introducing dropout \cite{srivastava2014dropout} for regularization will yield potentially significant improvements in test set accuracy on the virtual dataset. As an alternative regularization technique, we propose a virtual-real-world fusion data approach for training the models. Specifically, a batch of real-world data could be introduced every $N$ virtual-data batches. This approach will likely yield a considerable increase to real-world test set performance, as well as provide a feasible mechanism for bootstrapping real-world robotics systems that utilize deep learning methods for perception, planning, and navigation. In this paradigm of fusion training, only minimal real-world data would need to be collected, with the majority of the training coming from simulations. Conceptually, the models would learn rough approximations in the simulations, and refine important discriminating features via the interspersed real-world training batches.

Interestingly, the RNN outperformed the other models on the virtual dataset, and the evaluations of the models on the real-world dataset yielded counter-intuitive results. Predicated on a suspicion about the sequential order in which images are fed into the RNN, we ran a follow-up experiment wherein the RNN read the images from bottom-to-top as opposed to top-to-bottom. The performance of the RNN decreased substantially and rarely achieved greater accuracies than 50\% on the virtual test set. In general, the majority of salient features for trail perception are located within the bottom two-thirds of the image (i.e., the tips of trees is typically uninformative for discerning direction of a trail). When the image is fed from top-to-bottom into the RNN, the information in the top of the image is degraded due to vanishing gradients, which is a well established issue even for LSTM/GRU cell RNN's. Consequently, when images are fed in bottom-to-top, the most important information is now the first thing the RNN processes and is therefore mostly degraded from the recurrent connections by the end of the image feed.

This result is informative: it is a likely indicator that the classifier is learning to discriminate based on features within the lower half of the image. Consequently, computational demands may be lowered and training made more efficient by training on only bottom half or two-thirds of the image, reducing image processing time and decreasing the number of parameters in the model that must be trained.

To further understand the performance of our models, we analysed incorrectly classified images from the virtual test set. Our analysis points to deficiencies in the models when presented with multiple trails in a single image, suggesting the requirement of a higher level planning system---e.g., GPS and/or compass information of a goal position---to aid the robot's decision. This analysis also suggests that low quality terrain packs do not allow for sufficient variance amongst objects, obfuscating fine-grained distinctions between trails and other objects with similar features. Consequently, we believe the models may benefit from training on higher quality terrain packs. With state-of-the-art GPUs, virtual environments can be made to closely mimic the appearance of real-life environments and appear nearly photorealistic. We strongly believe virtual scene realism will play a direct role in transfer learning accuracy.

\subsection{Future Work}
Our model was trained using a virtual alpine environment and tested on real data of a similar terrain type. It is likely the model would perform much worse on environments that do not match the synthetic environment's general terrain characteristics and trail features. Future work will explore procedurally generating terrain with vastly different conditions and features (weather, lighting, biome, path appearance, elevation changes, flora, etc.) to improve generalizability while still being able to rapidly collect large synthetic training datasets. Conveniently, our work allows for rapidly exchanging terrain and environment packages, thus allowing for the development of navigation systems over a large variety of environments and conditions.

One major advantage of our approach is that our data collection process can be automated, drastically increasing the rate of labeled dataset generation. Our current approach captured 20,269 images in less than 5 minutes---a rate of 4,053 images per minute---and is in stark contrast to the 24,474 images collected over a period of 8 hours in \cite{giusti2016machine}. Future work may couple our automated data collection procedure with procedurally generated terrain with higher photorealism to produce additional improvements to this method.

Lastly, an interesting future direction is to discern which features are being learned in the classification task. In a virtual environment, over which we exert complete control, it is possible to filter out one feature at a time, and we can run the same classifier repeatedly in these slightly varied environments. If a feature is turned off and a significant perturbation to classification performance is measured, we can gain insight into the features important for the particular classification task. Running this experiment over numerous terrains may reveal globally important features, enabling us to leverage the statistical properties of these key features for procedurally generated terrain, optimizing the efficiency of the process and enabling more effective results.

\section{Conclusion}
In this paper, we trained three different neural network architectures on virtual data generated from Unity and achieved virtual-data classification accuracies ranging from 88\% to 95\% and real-world classification accuracies ranging from 38\% to 58\% over a baseline of 33.33\%. Robot battery life, human fatigue, and safety considerations present major challenges for manual data collection; however, with our approach, these issues may be circumvented as labeled data generation can be performed rapidly and efficiently within a virtual setting. Robots may then be virtually trained to navigate terrain that is hard to access and/or dangerous, including novel terrains that are currently impossible to access and collect real data from (e.g., Mars) without ever being first exposed to these environments. Our approach demonstrates that virtual-to-real-world transfer learning is a promising approach for overcoming the immense challenges facing real-world data collection and the development of autonomous robotics systems.

\addtolength{\textheight}{-12cm}   




\section*{ACKNOWLEDGMENTS}
This work was supported by a NSF CRII Award \#1566612 and an Early Career Faculty grant from
NASA's Space Technology Research Grants Program under award NNX16AR58G. We thank Michael C. Mozer for his help and support of this research.

\bibliographystyle{IEEEtranN}
\bibliography{unitydl}

\end{document}